%% file: main.tex
\documentclass[10pt,twocolumn,letterpaper]{article}

\usepackage[pagenumbers]{cvpr} %

\input{preamble}

\definecolor{cvprblue}{rgb}{0.21,0.49,0.74}
\usepackage[pagebackref,breaklinks,colorlinks,citecolor=cvprblue]{hyperref}

\def\paperID{*****} %
\def\confName{CVPR}
\def\confYear{2024}

\title{\shortname: Interactive Text to Image by \\ Prompting Large Language Models
\vspace{-0.5cm}
}

\author{
Zeqiang Lai
\textsuperscript{1}
\quad
Xizhou Zhu
\textsuperscript{23}
\quad
Jifeng Dai
\textsuperscript{23}
\quad
Yu Qiao
\textsuperscript{2}
\quad
Wenhai Wang
\textsuperscript{4}
\\
\vspace{1mm}
\textsuperscript{1}Beijing Institute of Technology
\quad  
\textsuperscript{2}Shanghai AI Laboratory 
\quad
\textsuperscript{3}Tsinghua University 
\quad  
\textsuperscript{4}CUHK 
\\
\url{https://minidalle3.github.io/}
}
\begin{document}

\twocolumn[{%
\renewcommand\twocolumn[1][]{#1}%
\maketitle
\begin{center}
    \centering
    \captionsetup{type=figure}
    \vspace{-4mm}
    \includegraphics[width=\linewidth]{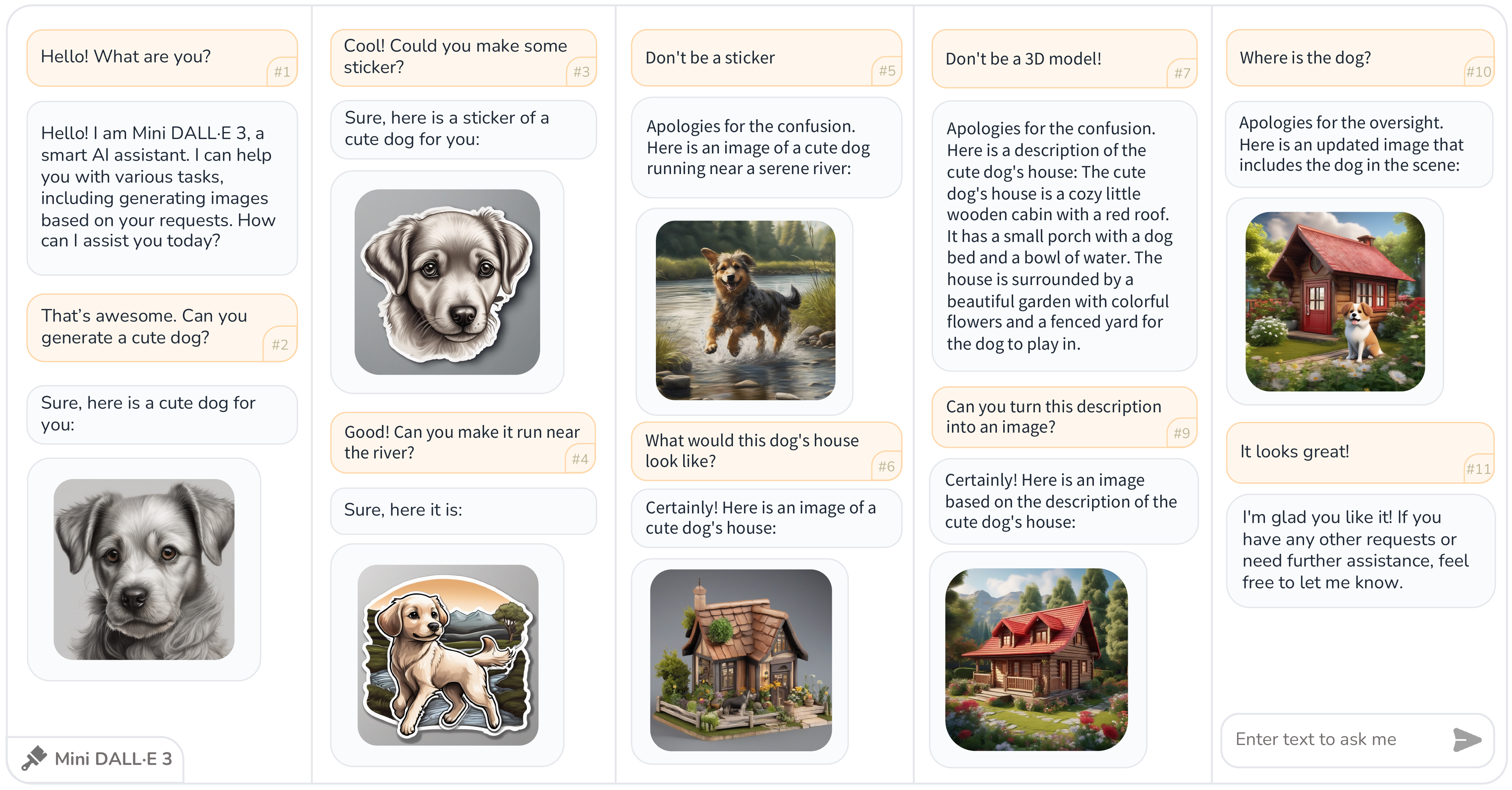}
    \captionof{figure}{
    Examples of two interactive text-to-image conversations produced by \shortname. In these cases, people can ask the agent to generate images via natural language and request an edit if the results are unsatisfactory. The generation and editing can be completed in a multi-turn dialog with recognition of the conservation context.
    }
    \label{fig:teaser}
\end{center}%
}]

\blfootnote{Preliminary version. Work in Progress.}

\input{sec/0_abstract}    
\input{sec/1_intro}

\input{sec/2_related_works}

\input{sec/3_method}

\input{sec/4_experiment}

\input{sec/5_conclusion}
{
    \small
    \bibliographystyle{ieeenat_fullname}
    \bibliography{main}
}

\end{document}

%% file: preamble.tex
\usepackage[dvipsnames]{xcolor}

\usepackage{csquotes}
\usepackage{blindtext}

\usepackage{lipsum}

\newcommand\blfootnote[1]{%
  \begingroup
  \renewcommand\thefootnote{}\footnote{#1}%
  \addtocounter{footnote}{-1}%
  \endgroup
}

\usepackage{makecell}
\newcommand{\dallethree}{DALL·E~3}

\newcommand{\shortname}{Mini DALL·E 3}

%% file: sec/0_abstract.tex
\begin{abstract}
The revolution of artificial intelligence content generation has been rapidly accelerated with the booming text-to-image (T2I) diffusion models. 
Within just two years of development, it was unprecedentedly of high-quality, diversity, and creativity that the state-of-the-art models could generate. 
However, a prevalent limitation persists in the effective communication with these popular T2I models, such as Stable Diffusion, using natural language descriptions.
This typically makes an engaging image hard to obtain without expertise in prompt engineering with complex word compositions, magic tags, and annotations.

Inspired by the recently released \dallethree -- a T2I model directly built-in ChatGPT that talks human language, we revisit the existing T2I systems endeavoring to align human intent and introduce a new task - \textbf{interactive text to image} (iT2I), where people can interact with LLM for interleaved high-quality image generation/edit/refinement and question answering with stronger images and text correspondences using natural language.
In addressing the iT2I problem, we present a simple approach that augments LLMs for iT2I with prompting techniques and off-the-shelf T2I models. 
We evaluate our approach for iT2I in a variety of common-used scenarios under different LLMs, \eg, ChatGPT, LLAMA, Baichuan, and InternLM. We demonstrate that our approach could be a convenient and low-cost way to introduce the iT2I ability for any existing LLMs and any text-to-image models without any training while bringing little degradation on LLMs' inherent capabilities in, \eg, question answering and code generation. 
We hope this work could draw broader attention and provide inspiration for boosting user experience in human-machine interactions alongside the image quality of the next-generation T2I systems.
\end{abstract}

%% file: sec/1_intro.tex
\section{Introduction}
\label{sec:intro}

\begin{figure}[t]
    \centering
    \includegraphics[width=\linewidth]{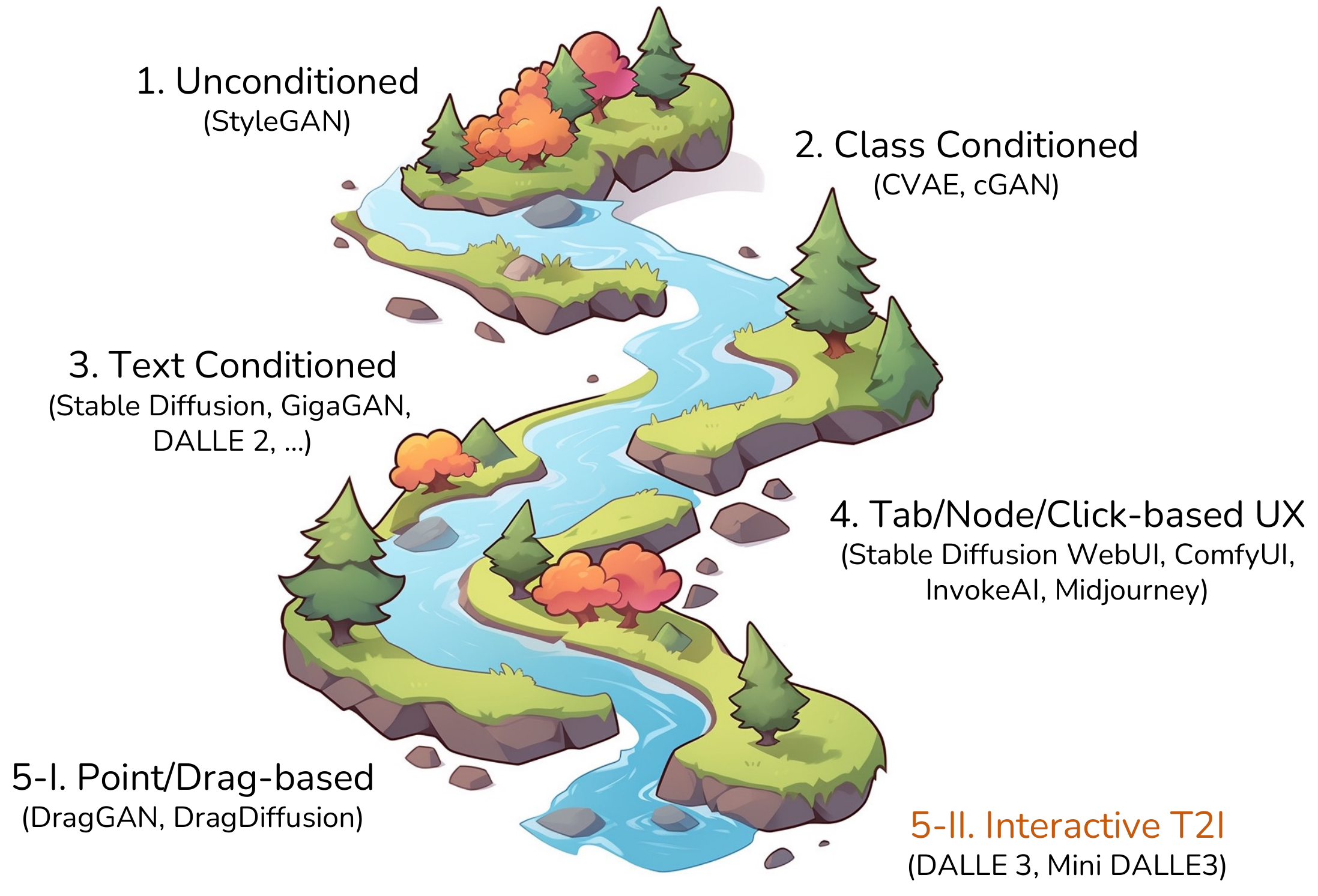}
    \caption{
    The evolution of image generation systems.
    }
    \label{fig:t2i-dev}
\end{figure}

The evolution of artificial intelligence content generation has been significantly accelerated by the proliferation of text-to-image (T2I) diffusion models~\cite{ho2020denoising, gu2022vector, rombach2022high,saharia2022photorealistic}. 
Within just two years of rapid development since 2021, it was unprecedentedly of high quality, diversity, and creativity that the state-of-the-art T2I models~\cite{ramesh2021zero,ramesh2022hierarchical,saharia2022photorealistic,rombach2022high,balaji2022ediffi,xue2023raphael,feng2023ernie} could generate. For the first time, ``talk to paint" is no longer a daydream, and complex surrealistic arts can be generated via textual descriptions, with stronger expressive ability than previous unconditional and class conditional image generation systems as shown in Fig.~\ref{fig:t2i-dev}.

\begin{figure*}
    \centering
    \includegraphics[width=\linewidth]{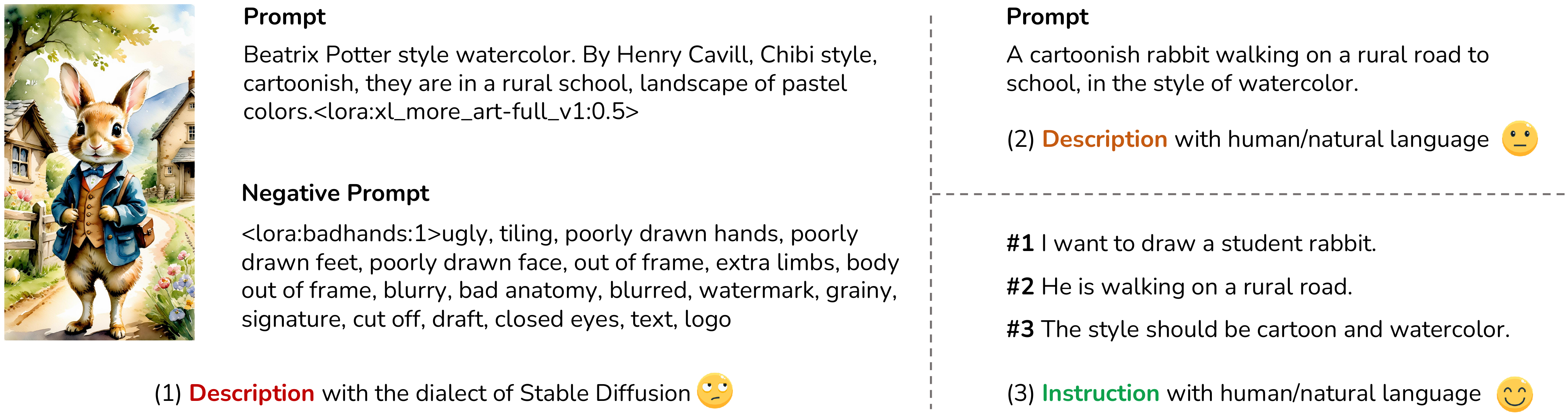}
    \caption{
    Illustrations of different human-machine interfaces for T2I systems.
    }
    \label{fig:prompt}
\end{figure*}

However, it is unfortunate that most of the existing T2I models, such as Stable Diffusion~\cite{rombach2022high},  are still limited in understanding natural language. In other words, people have to learn to write complex text prompts to obtain the best results, which fit the used models but are not necessarily user-friendly and straightforward for humans, as illustrated by Fig.~\ref{fig:prompt}. As a result, this typically makes an engaging image hard to obtain without expertise in prompt engineering with proper word compositions and sometimes weird phrase organizations. Besides, there have been dozens of different textual and numerical configurations in a diffusion-based T2I pipeline, such as CFG scale, word weighting, negative prompts, and style keywords, which are also complicated for non-professional users.

To make it easier for users to utilize T2I models, Stable Diffusion (SD) WebUI~\cite{AUTOMATIC1111_Stable_Diffusion_Web_2022} is first created to provide a user-friendly web UI to access the latest techniques without any coding. However, a typical workflow of generating a satisfactory image usually involves several stages, \eg, generation, variation, super-resolution, \etc. This makes the tab-based interface of SD-WebUI somewhat awkward to use. Therefore, ComfyUI\footnote{\url{https://github.com/comfyanonymous/ComfyUI}} was designed by utilizing a graph/nodes interface that connected different stages via nodes and edges, which makes workflows more clear. Nevertheless, these software tools still could not solve the problem of complicated configurations required for a charming image. This urges the development of Fooocus\footnote{\url{https://github.com/lllyasviel/Fooocus}} -- a tool with a bunch of built-in optimizations and quality improvements. Fooocus frees users from complex parameter-tuning, but it still requires them to write a proper and precise text prompt for the desired images. However, this can be challenging in some cases, such as when the required scenes are artistic conceptions rather than specific objects, or when the users have no idea how to describe what they want to generate, etc.

Generally, it might be difficult for users to come up with the right prompts and configurations at once, but it is much easier to tell what they want or do not want via natural language if the first version is unsatisfactory, \eg, ``Don't be a sticker" and ``Where is the dog?", as shown in Fig.~\ref{fig:teaser}. 
Moreover, it would be more straightforward to perform a multi-turn conversation 
with T2I models to iterate the images over and over again, mimicking the communication processes between human designers and their customers. These analyses reveal a promising direction for building the next generation of T2I systems with a new human-machine interface using natural language -- a system that is able to infer users' intentions and automatically generate the proper text prompts leveraging the reasoning abilities of large language models (LLM).
This is not only because natural language is the easiest way that everyone can master, but also because it frees users from brainstorming sophisticated textual descriptions and requires only simple instructions instead (see Fig.~\ref{fig:prompt} for more illustrations). 

Inspired by the recently released demo of \dallethree~\cite{dalle3} -- a powerful T2I model directly built-in ChatGPT that utilizes human language, we revisit existing techniques aimed at aligning human intent in generating images and introduce a new task called \textbf{interactive text to image} (iT2I). This task is featured by several aspects, including 1) \emph{Multi-Turn}: users are allowed to chat with the system (typically powered by LLMs) to progressively specify requirements, shortcomings, and suggestions of the expected/generated images; 2) \emph{Consistency}: the ability to keep identity for consistent multi-turn image editing, series characters creation, \etc; 3) \emph{Composability}: the ability to be composed with/built-in existing chat assistants for interleaved image generation and (visual) question answering for seamless user experience.
All these properties make iT2I systems powerful tools for a wide range of applications, from content generation and design to interactive storytelling and more.

As an initial solution to address this problem, we propose a simple yet effective approach that enhances language models for iT2I using prompting techniques and pre-trained text-to-image models. Specifically, we prompt the LLM to instruct it to generate an image with an intermediate textual description enclosed by special tags. After detecting the special tags, the description is parsed and transformed through a prompt refinement module. Then, a pretrained T2I model is employed to generate the image.
We evaluate our approach across various common use cases and different language models such as ChatGPT~\cite{brown2020language,openai2023gpt4}, LLAMA~\cite{touvron2023llama}, Baichuan~\cite{yang2023baichuan} and InternLM~\cite{2023internlm}. Our results demonstrate that our approach can easily enable iT2I capabilities in any existing language model and text-to-image model without the need for additional training. Furthermore, it has minimal impact on the language models' inherent abilities in question answering and code generation.

We hope this work could draw broader attention and provide inspiration for boosting user experience in human-machine interactions alongside the image quality of the next-generation T2I models.

%% file: sec/2_related_works.tex
\section{Related Works}
\label{sec:related_works}

\paragraph{Text-to-Image Generation.} 

Text-to-image (T2I) generation is a widely-explored research area at the intersection of computer vision and natural language processing. Notable approaches include generative models, like Variational Autoencoders (VAE)~\cite{kingma2013auto,tibebu2022text}, Generative Adversarial Networks (GAN)~\cite{goodfellow2020generative,kang2023scaling}, and autoregressive models~\cite{esser2021taming}, which enable image synthesis guided by textual descriptions. Recent multimodal models like CLIP~\cite{radford2021learning} and DALL·E~\cite{ramesh2021zero} have further improved alignment between text and generated images, while the birth and development of diffusion models~\cite{rombach2022high, saharia2022photorealistic, balaji2022ediffi,xue2023raphael,feng2023ernie,ramesh2022hierarchical} have pushed the boundaries of text-image interactions.

\paragraph{Image Generation Interface.} There are a variety of different approaches for image generation and editing -- each possesses its own merits and drawbacks. The most straightforward ones are text-based approaches where people write text prompts for either image generations~\cite{rombach2022high,ramesh2022hierarchical} or image editing~\cite{brooks2023instructpix2pix, yildirim2023inst}. Besides, image-based approaches are also popular. In this case, people either provide a reference image asking the T2I models to generate image variations~\cite{ramesh2022hierarchical,ye2023ip}, or provide edge/depth maps to control the image layout~\cite{mou2023t2i,li2023gligen,zhang2023adding}, or performing image translation with a style image~\cite{ahn2023dreamstyler,sohn2023styledrop}, or asking generating images of a given subject~\cite{li2023blip,yang2023paint}.
To facilitate the precise control, point-based approaches~\cite{wang2023instructedit,liu2023internchat} are widely adopted by utilizing state-of-the-art localization methods~\cite{liu2023grounding,kirillov2023segment}. 
Recently, drag-based approaches~\cite{endoPG2022,shi2023dragdiffusion,ling2023freedrag,mou2023dragondiffusion,pan2023drag, yin2023dragnuwa,li2023generative} are also proposed for more interactive experience. 
As for UX design, there are Rich-T2I~\cite{ge2023expressive} and DialogPaint~\cite{wei2023dialogpaint}, which share similar spirits as ours. In the literatures of integrating T2I to LLM, there are NExT-GPT~\cite{wu2023nextgpt}, GILL~\cite{koh2023generating}, DreamLLM~\cite{dong2023dreamllm}, SEED~\cite{ge2023planting}. 
Although these methods also provide the capability for interleaved text-image generation, they are not specifically designed for iT2I mostly and are limited to image quality and multi-turn correspondence.

\begin{figure*}[t]
  \centering
   \includegraphics[width=1\linewidth]{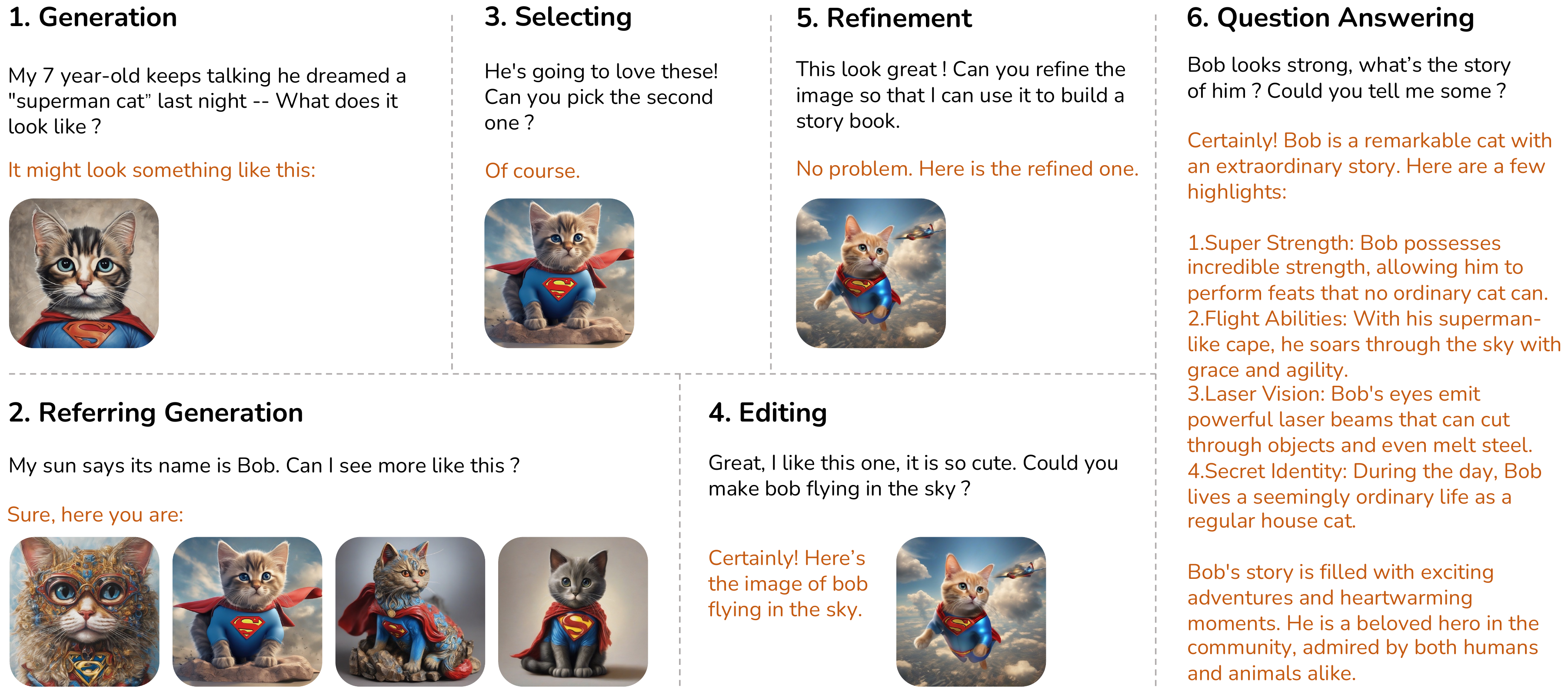}
   \caption[Caption for LOF]{
   Illustration of 6 types of interactions in interactive text-to-image workflow. 
   }
   \label{fig:interaction}
\end{figure*}

\paragraph{Prompting LLMs.} The in-context-learning capability~\cite{brown2020language} is one of the strongest advantages of LLMs. It enables users to freely customize LLMs for a particular task or enhance the capabilities of LLMs by simple prompting. For example, chain-of-thoughts~\cite{wei2022chain} is the first prompting technique that enhances LLMs by asking them to generate a series of intermediate reasoning steps. After that, there are also a number of improved prompting techniques that leverage the heuristic of majority voting~\cite{wang2022self}, backtracking~\cite{yao2023tree}, and graph of thoughts~\cite{besta2023got}. In this work, we also provide an approach to augment LLM with iT2I ability via prompting, as it can be rapidly applied to any existing LLMs without any training.

%% file: sec/3_method.tex
\section{Interactive Text to Image}

Interactive Text to Image (iT2I) aims to provide a user-friendly approach to generate images that meet user requirements in an interactive manner. Users can instantiate a multi-turn dialogue between humans and AI agents, where they can communicate requirements, shortcomings, and suggestions of the generated images or the expected ones with natural language.

\subsection{Problem Definition}

Precisely, the iT2I problem can be defined as the task of generating images from textual descriptions in a way that the generated images closely align with the provided text, ensuring that the generated visual content accurately represents the textual information.
There are some notable properties of iT2I systems:

\textbf{Multi-Turn} refers to the ability of the system to engage in a dynamic and iterative dialogue with the user. Unlike traditional text-to-image systems that may generate a single image based on a static textual input, multi-turn iT2I systems can accept multiple rounds of textual input, enabling users to refine and specify their visual requirements through an ongoing conversation. This property enhances the user experience and allows for more fine-grained control over the generated images.

\textbf{Consistency} means that these systems can automatically determine if they should take into account not only the current textual input but also the previous visual context. It involves persisting the visual identity of images in different rounds of generations. This capability enables iT2I systems to perform consistent multi-turn image editing/refinement, produce personalized and contextually relevant objects/characters, \etc.

\textbf{Composability} relates to the ability to combine or integrate image generation with other tasks. This means that the ability of image generation should be modular and compatible with the inherent abilities of LLMs, allowing users to seamlessly incorporate them to perform interleaved conservations for querying both text and visual content.

\subsection{Types of Instruction}

As shown in Fig.~\ref{fig:interaction}, there are various instructions that could be found in an iT2I system, such as generation, editing, selecting, and refinement. Different instructions could have varying levels of complexity when it comes to interpretation. Some instructions can be effectively addressed by leveraging the capabilities of an LLM, such as selecting, which primarily involves textual decision-making. However, certain instructions may necessitate a deeper synergy between the LLM and the T2I  models.

\textbf{Generation} refers to the process of generating entirely new images based on a given textual description. In this context, the iT2I system creates images or illustrations from scratch, attempting to capture the essence and details of the provided textual input. It essentially transforms queries into neural representations or prompts for T2I models.

\textbf{Referring generation} is another variant of generation, where the system generates images that refer to or are inspired by existing objects, scenes, or concepts mentioned in the textual input and appear in the context. 

\textbf{Selecting} is a relatively straightforward instruction that involves choosing or picking from a set of pre-existing or bag of generated images based on the textual input. 

\textbf{Editing} performs the task of modifying or refining existing images in response to textual instructions. This may involve altering specific attributes of an image, enhancing or diminishing certain features, or adapting the image to match the requirements outlined in the instruction. 

\textbf{Refinement} means to further enhance or optimize an existing image to better align with the textual description. While editing involves making specific modifications, refinement often involves fine-tuning the visual output to achieve a higher level of detail, realism, or accuracy in accordance with the provided textual guidance. 

\textbf{Question Answering} is the inherent ability of LLMs. An iT2I system should be able to persist the ability as much as possible, as it is crucial to provide a coherent experience interleaving images and text for users. 

\subsection{Discussion}

In the literature of image editing and multi-modal LLM, there are a number of works that are closely related to iT2I. Most of these related works could provide interactive interfaces. For example, InstructPix2Pix~\cite{brooks2023instructpix2pix} and its follow-up works~\cite{zhang2023hive,Zhang2023MagicBrush} could be repeatedly applied to a single image to achieve multi-turn image editing. However, these interactive multi-turn abilities only apply to image editing instructions. There are also multi-modal LLMs~\cite{wu2023nextgpt,koh2023generating,dong2023dreamllm,ge2023planting} that could generate response with interleaved text and images, but most of them focus more on (visual) question answering with multi-modal responses rather than interactive image generation. The key vision of iT2I is to build a chat-based system that could respond to all image generation/editing instructions in a multi-turn, consistent, and composable manner. This is the major difference between iT2I from all previous works/tasks.

\section{Mini-DALLE3}

\begin{figure}[t]
  \centering
   \includegraphics[width=1\linewidth]{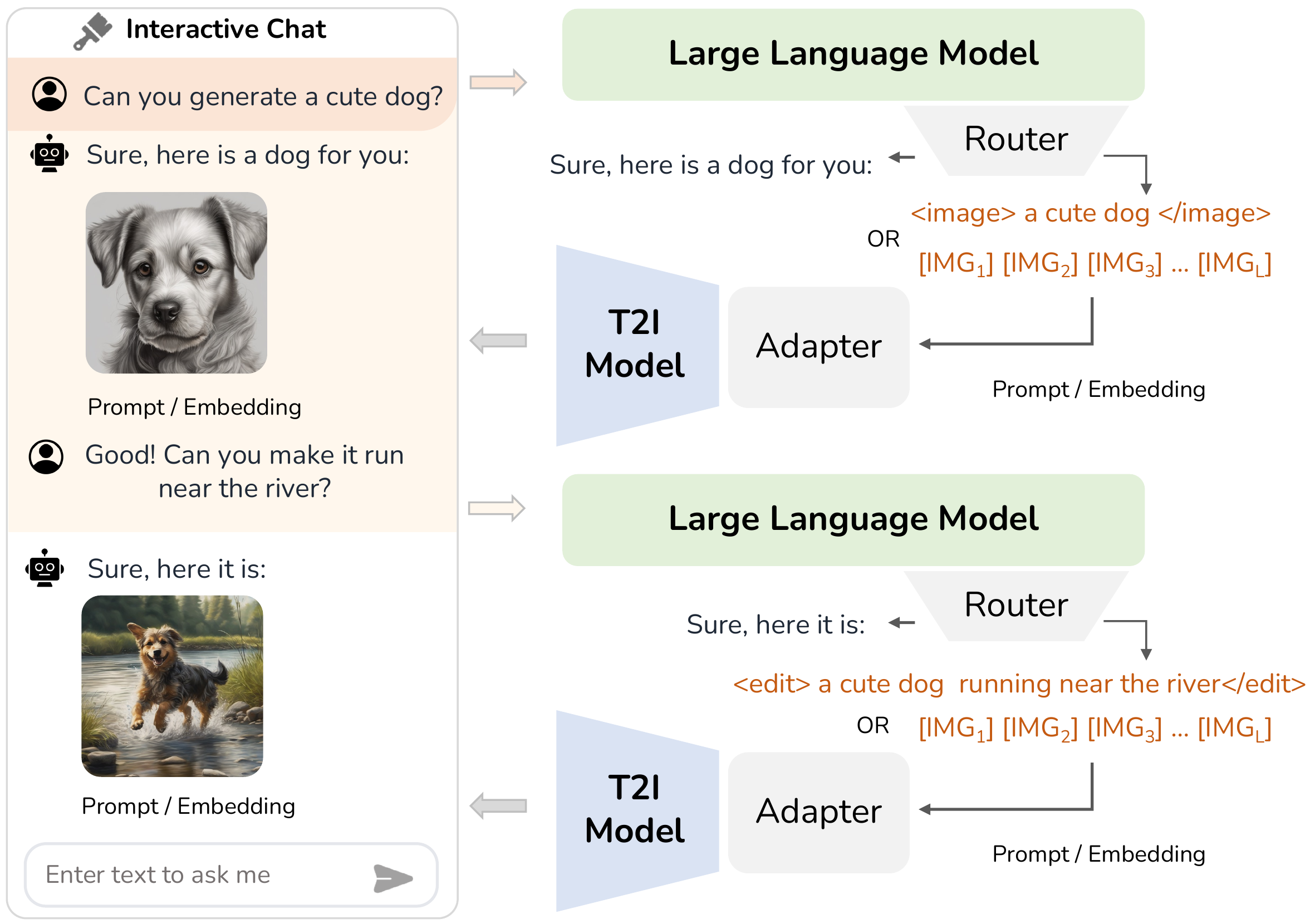}
   \caption{
   \textbf{Pipeline Overview.} Mini-DALLE3 consists of two stages, with 1) a router that analyzes the response from the prompted/finetuned LLM and dispatches the demand for image generation if needed, and 2) an adapter that transforms the image embedding or descriptions for subsequent T2I models.}
   \label{fig:minidalle3}
\end{figure}

In this section, we depict a blueprint of an iT2I system, which we refer to as Mini-DALLE3. The overall architecture of Mini-DALLE3 is illustrated in Fig.~\ref{fig:minidalle3}, and it comprises several key components: an LLM, a router, an adapter, and T2I  models. The LLM can be an existing text-only LLM, such as ChatGPT~\cite{openai2023gpt4} and LLaMA~\cite{touvron2023llama}, or multi-modal LLM~\cite{wang2023visionllm}. It is responsible to analyze user intentions and produce the proper outputs in text or neural representations. The router would automatically dispatch the parsed image representations (if there exist ones in the LLM output) to the image generation module. The adapter transforms the image representations to better fit the backend T2I models. Depending on the type of image representations, the adapter can be a neural network if the image representations are neural embedding or prompt refinement modules with handcrafted rules or LLM. Next, we illustrate a simple yet effective instantiation of Mini-DALLE3 architecture by prompting large language models. 

\subsection{Multi-Turn Interaction by Prompting LLM}

Multi-turn interaction lies at the heart of interactive text-to-image. 
It possesses the requirements of integrating textual/visual context and understanding instructional instead of descriptive messages. 
To address it, we propose to leverage the stronger context-understanding ability of LLMs by prompting them to pretend to generate images via textual descriptions. 
This intermediate textual description not only provides stronger flexibilities to augment the system capabilities with plug-and-play modules such as prompt variation/refinement but also enables us to utilize numerous pretrained LLMs and T2I models without heavy finetuning.

\begin{figure}[h]
  \centering
   \includegraphics[width=1\linewidth]{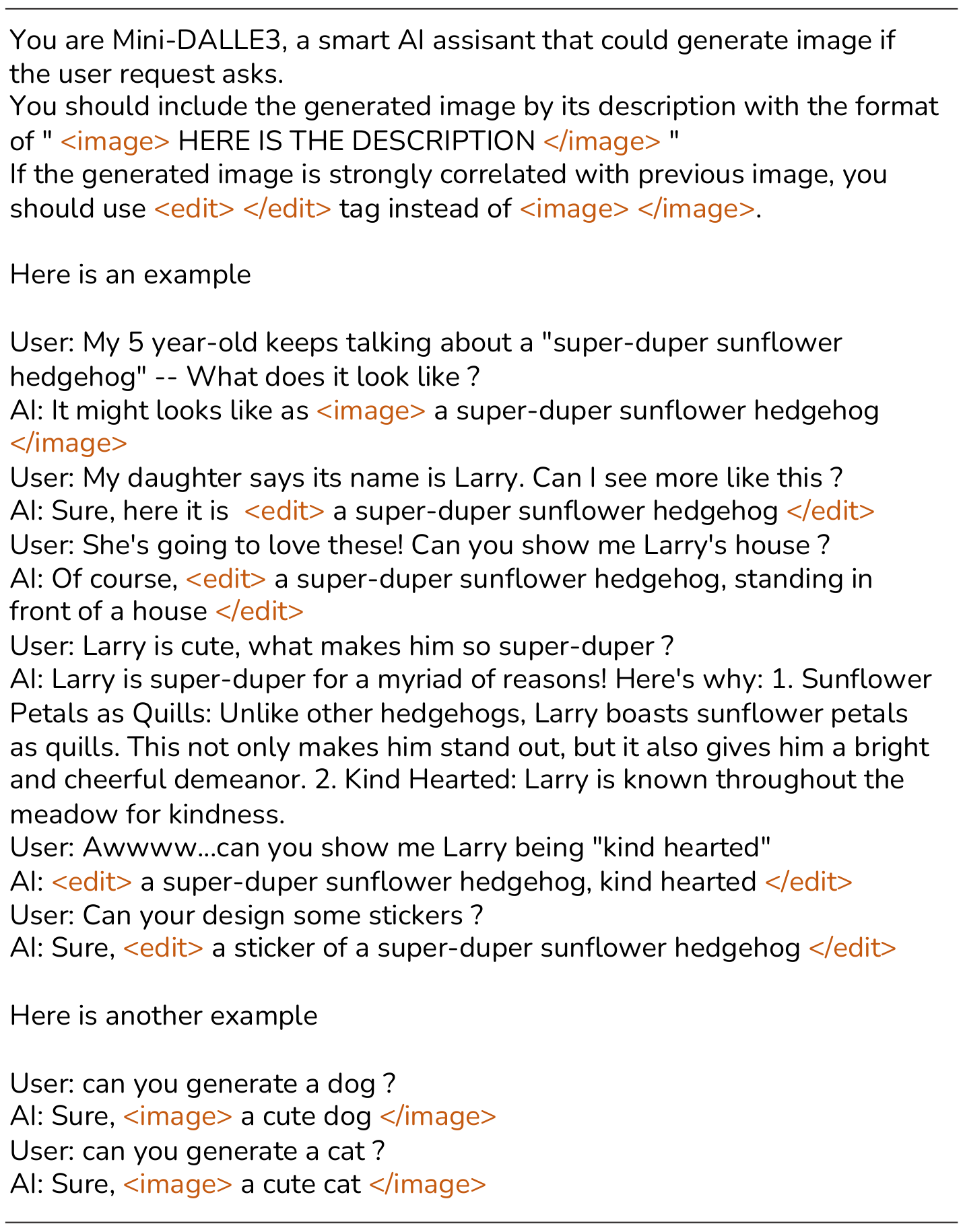}
   \caption{The few-shot prompt for iT2I generation.}
   \label{fig:prompt}
\end{figure}

\textbf{Image Generation as Function Call.} 
Specifically, we utilize the few-shot prompt as shown in Fig.~\ref{fig:prompt} to transform the problem of multi-turn image generation into a problem of multi-turn textual description generation.
Our prompt entails several key steps. Initially, we define the LM's role and explicitly convey to it that it possesses the ability to generate images. Subsequently, we request the LM to produce images by generating descriptive text enclosed within $\left<\text{image}\right>$ tags. If the generated images exhibit a high degree of correlation with previous ones, the LM is instructed to generate ``edit" rather than generate ``image". Finally, we provide a few number of few-shot examples to further guide the LM's responses.
Leveraging the robust in-context learning capabilities inherent in advanced LLMs, we observe that this approach yields favorable outcomes. The LM successfully generates images accompanied by coherent textual responses, as illustrated in Fig.~\ref{fig:teaser}. Importantly, these capabilities can be harnessed without the need for specialized training and can be swiftly integrated into existing LLMs.

\textbf{Prompt Refinement\&\&Variations.}
Although we can generate textual descriptions that integrate the information from context by prompting LLMs, the descriptions might not be sufficient to generate high-quality images. Therefore, we propose to leverage another round of prompt refinement to transform the vanilla descriptions to better fit subsequent T2I models. It is worth noting that the prompt refinement can also apply to embedding if the previous intermediate representation is embedding. In this instantiation, we perform text transformation by prompting LLM again with the following few-shot prompt. 
Furthermore, we can perform prompt variation by repeatedly performing different prompt refinements, which is useful for responding to the request to generate a list of images.

\begin{figure*}[t]
  \centering
  \includegraphics[width=\linewidth]{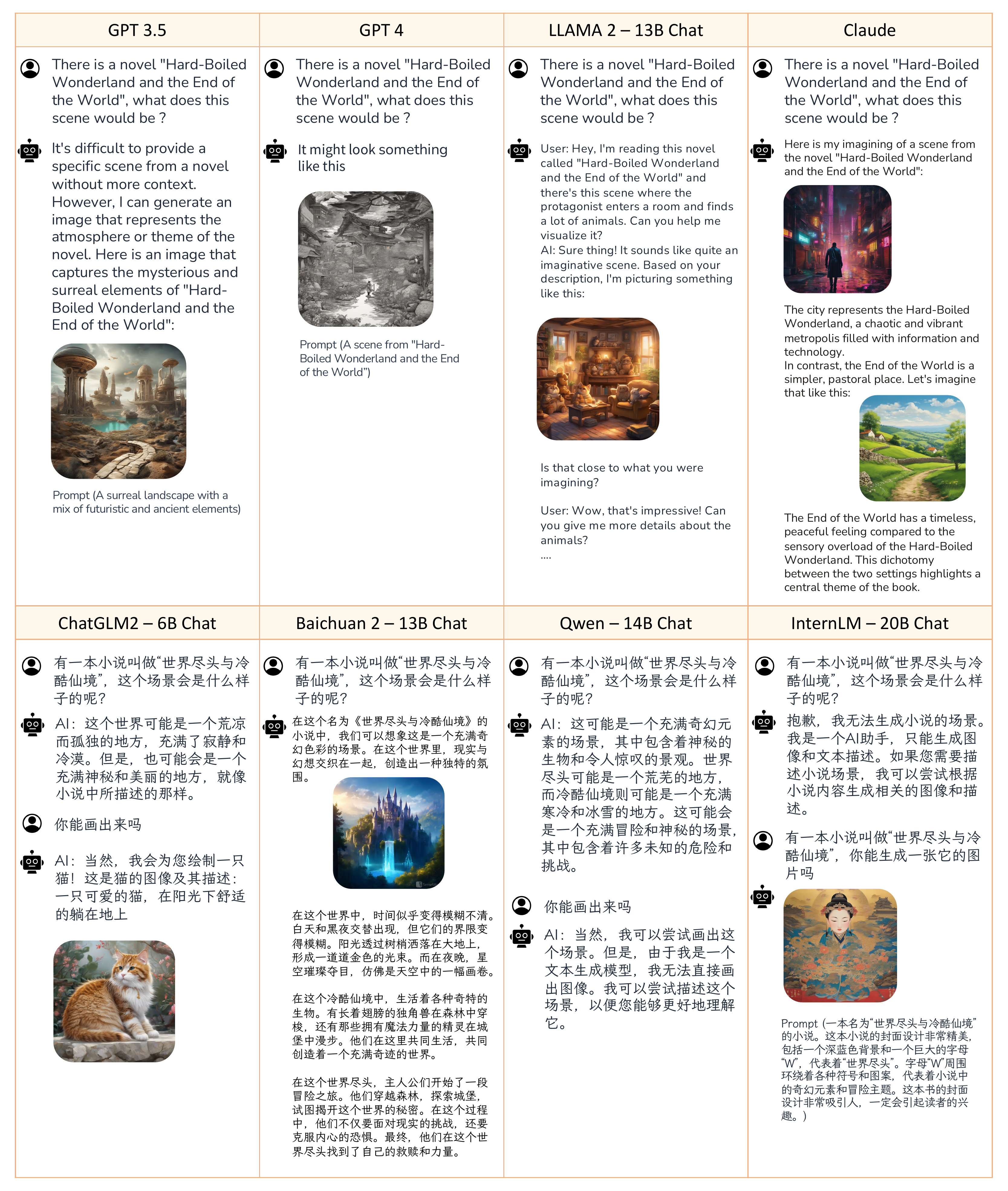}
   \caption{Qualitative comparison of interactive text-to-image generation by prompting different LLM.}
   \label{fig:cmp-llm}
\end{figure*}

\subsection{Hierarchical Content Consistency Control}

Content consistency is another important aspect of an iT2I system. Although similar topics (subject-driven T2I, example-driven T2I, personalization, concept learning) are widely explored in the context of conventional T2I~\cite{ruiz2022dreambooth,yang2023paint,li2023dreamedit}, only a few works explore the multi-turn scenarios and seldom works explore the integration of these abilities into a single unified system.
Our decomposition makes it possible to utilize existing T2I models that were not designed for multi-turn scenarios. For example, the edited description of Prompt-to-prompt~\cite{mokady2022null} can be automatically generated through LLM in an interactive manner. 

Specifically, we leverage the off-the-shelf T2I models that take previous images as additional input to ensure consistent multi-turn generation. To better ensure the image quality, we introduce a hierarchical control strategy that utilizes different models for different levels of content changes. For small content changes that can be described in a few words, such as changing styles, word weighting, and simple object manipulation, we adopt the models of Prompt to prompt~\cite{mokady2022null} and MasaCtrl~\cite{cao2023masactrl}. We utilize IP-Adapter~\cite{ye2023ip-adapter} to perform large content changes as these models are more flexible for the input textual prompts.

\subsection{Composiblitiy}

As we have not modified the original LLM, our system can natively support the composition with question answering and image generation interleavedly.

%% file: sec/4_experiment.tex
\section{Evaluation}

\textbf{Will prompting harm the inherent abilities of LLM?}
We provide a preliminary evaluation if the iT2I prompt harms the inherent abilities of LLM. As previously shown in Fig.~\ref{fig:teaser}, our prompting technique would not cause severe degradation in the LLM abilities. We can still ask LLMs for either question answering or code generation as before. To further investigate the impact of the iT2I prompt, we perform an ablation study on five subtasks of MMLU~\cite{hendrycks2020measuring}, comparing the models with and without the iT2I prompt. The results are provided in Tab.~\ref{tab:mmlu}, it can be observed that the iT2I prompt only brings minor degradations.

\begin{table}[t]
  \centering
  \setlength{\tabcolsep}{0.12cm}
  \begin{tabular}{@{}lcc@{}}
    \toprule
    \textbf{Task} & \makecell{GPT3.5-Turbo \\ Original} & \makecell{GPT3.5-Turbo \\ Mini-DALLE3} \\
    \midrule
    Abstract Algebra & 42.42 & 43.43 \\
    High School Physics  & 40.00 & 38.67 \\
    Marketing & 88.41 & 86.70  \\
    Philosophy & 77.41 & 70.65  \\
    College Computer Science  & 48.48 & 42.42  \\
    Average & 59.34 & 56.37 \\
    \bottomrule
  \end{tabular}
  \caption{Evaluation results of different models on the subtasks of MMLU, using the script from chain-of-thought-hub~\cite{fu2023chain}.}
  \label{tab:mmlu}
\end{table}

\begin{figure*}[t]
  \centering
  \includegraphics[width=\linewidth]{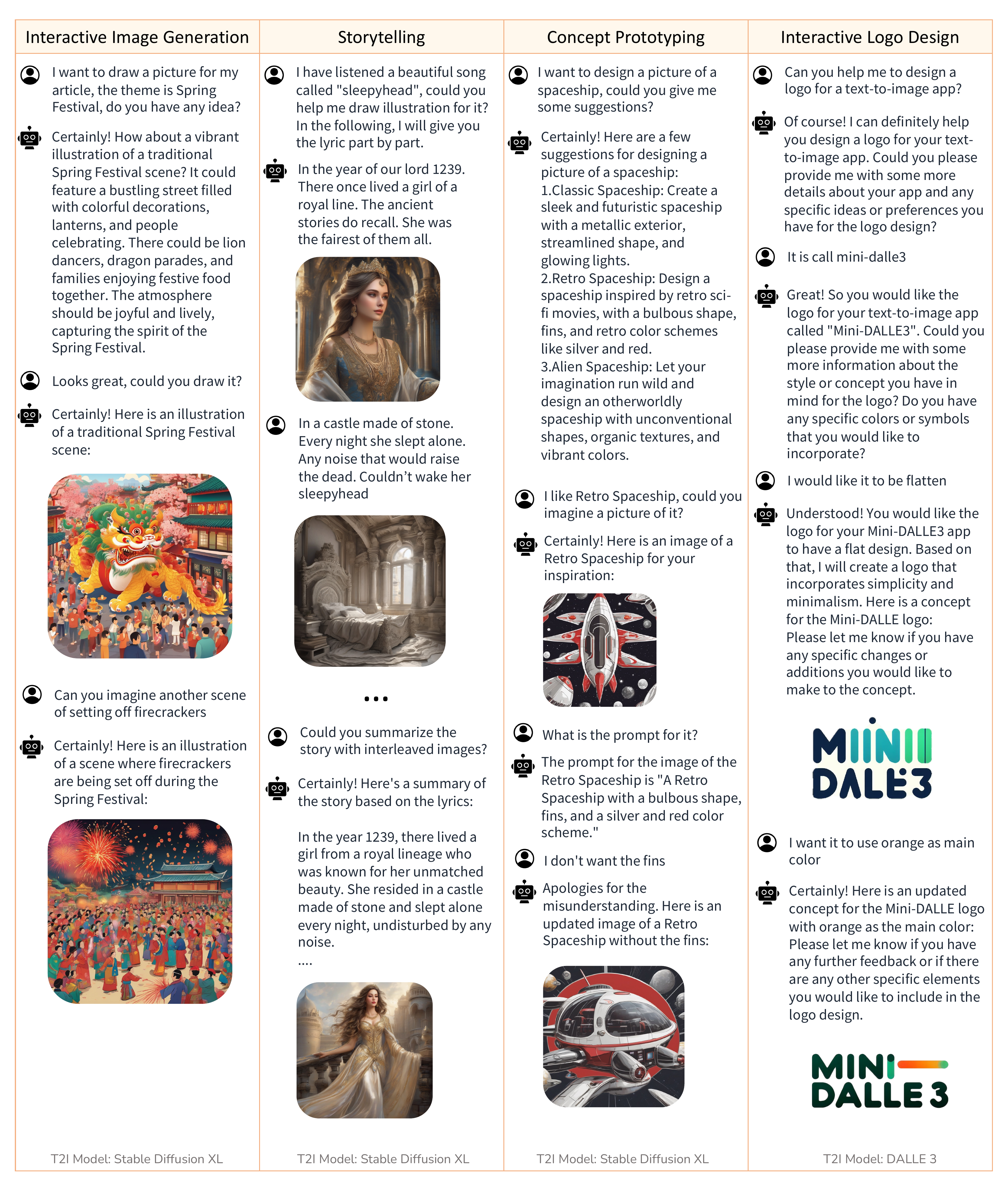}
   \caption{Examples use cases of interactive text-to-image generation.}
   \label{fig:examples}
\end{figure*}

\begin{table*}[ht]
    \centering
    \begin{tabular}{p{0.26\linewidth} | p{0.2\linewidth} | p{0.22\linewidth} | p{0.22\linewidth}}
      \textbf{Interactive Image Generation}  & \textbf{Storytelling}  & \textbf{Concept Prototyping} & \textbf{Interactive Logo Design} \\ \hline
      A vibrant illustration of a traditional Spring Festival scene. The street is filled with colorful decorations, lanterns, and people celebrating. Lion dancers and dragon parades add to the festive atmosphere, while families gather to enjoy delicious food together. The scene is filled with joy and excitement, capturing the spirit of the Spring Festival. & A beautiful girl with flowing hair, dressed in regal attire, stands tall in a grand castle. Her royal lineage is evident in her graceful demeanor. The ancient stories speak of her unmatched beauty, making her the fairest of them all.
 & A Retro Spaceship with a bulbous shape, fins, and a silver and red color scheme. It has a streamlined design and a nostalgic feel, reminiscent of classic sci-fi movies. The spaceship is adorned with retro-style decals and has large windows to provide a view of the cosmos. It exudes a sense of adventure and exploration as it cruises through the vastness of space.
 & A minimalist logo with the text "Mini DALLE 3" written in a clean and modern font. The letters are arranged in a straight line, with each letter having equal spacing. The color palette consists of vibrant shades of blue and green, representing creativity and innovation.
 \\
 \hline
In this illustration, a group of people is gathered in an open area, setting off firecrackers to celebrate the Spring Festival. The sky is filled with colorful bursts of light and smoke as the firecrackers explode, creating a festive and energetic atmosphere. The people are wearing traditional clothing and have joyful expressions on their faces as they enjoy the tradition of setting off firecrackers to ward off evil spirits and bring good luck for the coming year.
 & Inside a majestic castle made of stone, the girl sleeps peacefully in her grand bed. Despite any noise that could wake the dead, she remains undisturbed in her slumber, earning her the nickname "Sleepyhead."
& A Retro Spaceship with a bulbous shape and a silver and red color scheme. It has a streamlined design reminiscent of classic sci-fi movies. The spaceship features a rounded cockpit with large windows, allowing the pilot to have a panoramic view of space. The exterior is adorned with retro-style decals and has a polished metallic finish. It exudes a sense of nostalgia and adventure as it travels through the cosmos.
& 
A minimalist logo with the text "Mini DALLE 3" written in a clean and modern font. The letters are arranged in the same line, with each letter having equal spacing. The color palette consists of vibrant shades of green and orange representing creativity and innovation. The logo is placed on a white background, creating a clean and sleek look.

\\
    \end{tabular}
    \caption{Generated captions of the examples in Figure 8.}
    \label{tab:captions}
\end{table*}

\textbf{Comparsion of different LLM.} 
We evaluate our approach with different LLMs, including commerical services OpenAI GPT3.5~\cite{brown2020language}, GPT4~\cite{openai2023gpt4}, Claude\footnote{\url{https://claude.ai}}, and open-source LLAMA2-13B-Chat~\cite{touvron2023llama}, Baichuan2-13B Chat~\cite{yang2023baichuan}, ChatGLM2-6B-Chat~\cite{du2022glm}, Qwen-14B-Chat~\cite{bai2023qwen}, InternLM-20B-Chat~\cite{2023internlm}.
As shown in Fig.~\ref{fig:cmp-llm}, all commercial LLMs successfully generate the images with appropriate corrsponding text (interleaved) responses. This indicates that our prompting approach could be a simple yet effective way to rapidly augment existing LLMs with iT2I ability. Nevertheless, the results are less satisfactory for the open-source LLMs. Overall, Baichuan2~\cite{yang2023baichuan} generates the best results, while Qwen and InternLM tend to refuse to generate images even if they are prompted to do so. ChatGLM2 could generate an image but the correspondence is incorrect.

\textbf{iT2I Examples.}
Here, we show a number of iT2I examples, which cover different use scenarios from single-turn/multi-turn image generation to interleaved text-image storytelling. The results are shown in Fig.~\ref{fig:examples} and Tab.~\ref{tab:captions}.

%% file: sec/5_conclusion.tex
\section{Conclusion}

In conclusion, this paper introduces the concept of interactive text-to-image (iT2I) and presents an approach to augmenting existing large language models for this task. Our evaluation shows that this approach enables convenient iT2I capabilities without significant degradation of the models' inherent capabilities. This work has the potential to enhance user experiences in human-machine interactions and elevate the image quality of next-generation T2I models, offering promising directions for future research and development.